**Data Mining Project Report**
Pace University, May 10, 2023

# *Galaxy Classification: A machine learning approach for classifying shapes using numerical data*

Anusha Guruprasad

*Department of Computer Science Pace University*
New York, USA

ag84104n@pace.edu

## *Abstract*

*The classification of galaxies as spirals or ellipticals is a crucial task in understanding their formation and evolution. With the arrival of large-scale astronomical surveys, such as the Sloan Digital Sky Survey (SDSS), astronomers now have access to images of a vast number of galaxies. However, the visual inspection of these images is an impossible task for humans due to the sheer number of galaxies to be analyzed. To solve this problem, the Galaxy Zoo project was created to engage thousands of citizen scientists to classify the galaxies based on their visual features. In this paper, we present a machine learning model for galaxy classification using numerical data from the Galaxy Zoo[5] project. Our model utilizes a convolutional neural network architecture to extract features from galaxy images and classify them into spirals or ellipticals. We demonstrate the effectiveness of our model by comparing its performance with that of human classifiers using a subset of the Galaxy Zoo dataset. Our results show that our model achieves high accuracy in classifying galaxies and has the potential to significantly enhance our understanding of the formation and evolution of galaxies.*

## 1. Introduction

Galaxies have fascinated astronomers and scientists for centuries. The study of galaxies has come a long way since the days of Aristotle, who believed that the Milky Way galaxy was the entire universe. Over time, observations and advances in technology have revealed that there are millions of other galaxies in the universe, each with its unique characteristics.

Galaxies are classified into three main types: elliptical, spiral, and uncertain. Elliptical galaxies are oval or round, with stars that are evenly speckled throughout. Spiral galaxies have a distinctive spiral shape with extending "arms" and a flat disk that has a bulge in the center. Uncertain galaxies have no specific shape and are often chaotic in appearance.

The term "galaxy" itself has an interesting history. It comes from the Greek word "galaxias," meaning "milky," which is a reference to the Milky Way galaxy. The Milky Way was once thought to be the center of the universe, but this idea was challenged in the early 1900s by Harlow Shapley, who identified spiral-shaped blobs as separate from the Milky Way galaxy. As observations and classification systems improved, scientists have continued to refine their understanding of galaxy types and characteristics.

Edwin Hubble built upon Shapley's work by identifying pulsing stars called "Cepheid variables" in 1924. He realized that these stars were outside of the Milky Way and were a unique set of stars that existed at far away distances. Hubble created the first classification system for galaxies based on his observations and Gerard de Vaucouleurs later revised it to identify the three main galaxy types: elliptical, spiral, and uncertain. De Vaucouleurs broke these types down by characteristics such as openness of spirals, extent and size of bars, and galactic bulge size. As scientists continued to observe galaxies, they added additional sub-classifications that included markers like the star-formation rate of a galaxy and the age spectrum of the stars in a galaxy.

In this paper, we will explore the history of galaxies, their types, and the growth of the term "galaxy" while focusing on the application of machine learning to classify galaxy shapes. The goal of this study is to develop a more efficient and accurate method for classifying galaxies based on their numerical data.

## 2. Input and Output

The Galaxy Zoo dataset obtained from the SDSS Data Release 7 includes classifications of galaxies. The dataset contains the following columns:

### A. Input

| ATTRIBUTE | DESCRIPTION | FORMAT |
|---|---|---|
| Object ID | Unique identifier for each galaxy | Integer |
| Spectra | Indicates whether the spectra of the galaxy are included. | Float |
| Vote Fraction | The fraction of votes received in each of the six categories. | Float |
| Debiased Vote Elliptical | Debiased votes specifically for the elliptical category. | Float |
| Debiased Vote Spiral: | Debiased votes specifically for the spiral category | Float |
| Flag Spiral | Flag indicating whether the system is classified as a spiral galaxy. | Boolean |
| Flag Elliptical | Flag indicating whether the system is classified as an elliptical galaxy. | Boolean |
| Flag Uncertain | Flag indicating whether the system is classified as uncertain. | Boolean |

### B. Output

The target variable for classification is the type of galaxy. The dataset includes the following columns for the target variable:

| ATTRIBUTE | DESCRIPTION | FORMAT |
|---|---|---|
| Shape_of_galaxy | The type of galaxy assigned specific value (e.g., 0 for Spiral, 1 for Elliptical, and 2 for Uncertain). These values are assigned to represent the respective types of galaxies | Integer |

*Note: The individual columns for Spiral, Elliptical, and Uncertain types of galaxies have been dropped, as they are now represented by the Shape_of_galaxy column.*

*The dataset can be used for training classification models to predict the type of galaxy based on the provided features and target variable.*

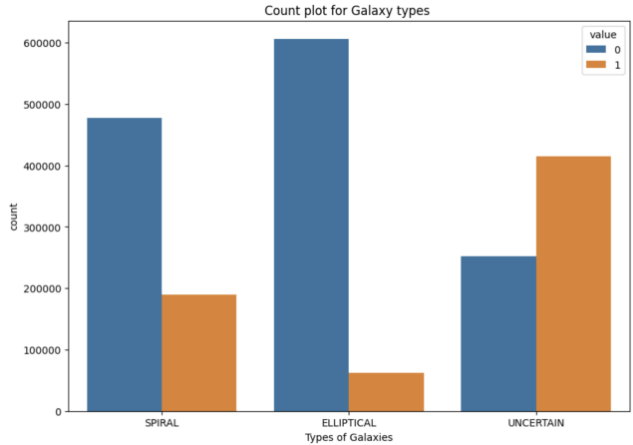

Fig 2.1 Output data distribution

## 3. Algorithms

### 3.1 K Nearest Neighbor for Galaxy Classification

KNN is a non-parametric supervised learning algorithm. Its most common method in pattern recognition has many applications in clustering and classification problems[8]. KNN tends to find similar data; feature space between data is defined by the similarity in distance. KNN can predict the type of galaxies that is spherical, elliptical, or uncertain, comparing the neighboring galaxies in feature space the algorithm can determine the similarity between the galaxies.

KNN is a simple algorithm for predicting the class of observations from their nearest neighbors in a feature space. Similarity in distance between the data points defines the feature space. It is necessary to set the number of neighbors to consider, K, before running the algorithm.

In KNN there are many ways to calculate distance, but the most common method is Euclidean distance (L2), the algorithm stores, the method simply keeps all the training instances (x, f (x)) in memory, where x is an n-dimensional feature vector (a1, a2, a3...an) and f (x) is the matching output[2]. Beyond these training examples, generalization is postponed until a new instance must be classified. Given a query xq, KNN finds the k training examples that are most like it (its k nearest neighbors) using the standard Euclidean distance as a measure of similarity between each training example xi and the query point xq: where $a_r$ is the value of the rth attribute of the instance x[2].

$$d(x_q, x_i) = \sqrt{\sum_{r=1}^{n}(a_r(x_q) - a_r(x_i))^2},$$

KNN delivers the most common target function value among the query point's neighbors when the target function is discrete-valued. The KNN technique gathers the nearest k neighbors and allows them to vote on which class wins[1]. The parameter k of the k-nearest neighbors algorithm should be chosen between 1 and the total number of samples. Higher k values give smoothing, which minimizes the susceptibility to noise in the training data. It is worth noting that if k is selected as the total number of samples in the training set, then for every new instance, all the examples in the training set become the nearest neighbors. In this situation, the anticipated answer for each new test case is simply the most common. For our dataset we have taken 10 features and selected the appropriate value for k which represents the number of nearest neighbors for classification.

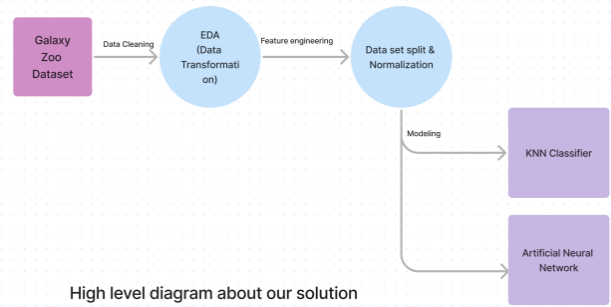

Fig 3.1.1 High level diagram

## A. Algorithm

1. Input, splitting the dataset into test and train data. Training is used for training the KNN model and test is used for evaluating the performance of the model.
2. Selecting the K value for appropriate value for the nearest neighbor classification and evaluating distance metric like Euclidean distance to measure the similarity between galaxies.
3. Training the KNN model
4. Calculate the distances to all instances in the training set based on the categorization of each instance in the testing set. Determine the K closest neighbors based on the smallest distance values.
5. Determine the majority class among the K nearest neighbors and apply it to the test instance as the anticipated class.
6. To evaluate the accuracy and performance of the KNN model, compare the predicted classes with the actual labels in the testing set.

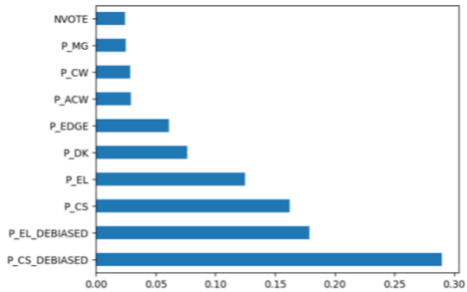

Fig 3.1.2 Features used in KNN model showing feature importance

## B. Implementation

The KNN classification model is a robust method that is trained with a specified parameter set, in this case, 5 closest neighbors. These neighbors are employed to generate predictions on both the training and testing datasets. We can examine the model's accuracy by comparing the predicted labels to the actual labels in these datasets, which offers useful insights into its classification performance.

A pipeline technique is used to further increase the accuracy and performance of the KNN model. This pipeline is made up of three major components: a KNeighborsTransformer, a KNeighborsClassifier, and GridSearchCV. The KNeighborsTransformer computes the nearest neighbors graph by employing the maximum number of neighbors necessary in the subsequent grid search. The KNeighborsClassifier, which filters the nearest neighbors graph based on its own n_neighbors parameter, is then used to do the classification operation. GridSearchCV performs an exhaustive search across a given parameter grid, enabling us to experiment with alternative values for the n_neighbors parameter[9].

A pipeline and grid search are used to identify the optimal n_neighbors value that yields the best KNN classifier accuracy. Using GridSearchCV, we can systematically search for the best-performing model by evaluating multiple combinations of n_neighbors. Grid search results are visualized in a plot, which shows the classification accuracy achieved with various n_neighbor values. By analyzing the plot, we can determine which value maximizes the accuracy score. The best accuracy achieved is 89% in this case.

Overall, the use of the KNN algorithm, pipeline, and grid search helps us to improve the classification model's performance. This provides us with useful information about the categorization problem at hand. Increasing the model's accuracy and reliability will allow us to generate more dependable predictions in a range of real-world settings.

| Metric | Accuracy |
|---|---|
| Training Accuracy | 0.92092 |
| Test Accuracy | 0.887648 |
| Best Accuracy | 0.890816 |

Fig. 3.1.2 Accuracies over datasets and best accuracy determined

## C. Outcome:

We achieved an accuracy of 89% on the classification test by using the KNN classification model with 5 closest neighbors and using the pipeline and grid search approaches. This high accuracy indicates the KNN algorithm's efficiency in reliably predicting the labels of cases in the dataset. The grid search analysis enabled us to determine the best value for the n_neighbors parameter, which considerably improved the model's performance.

Using the KNN algorithm for classification tasks has been shown to have valuable outcomes in this study. In combination with the grid search, the pipeline approach allows us to fine-tune the model and select the best configuration. Based on the 89% accuracy achieved by the model, it can successfully classify instances and make reliable predictions[9].

We can improve the accuracy and reliability of our models by understanding the strengths and limitations of the KNN algorithm and optimizing its parameters through pipeline and grid search techniques.

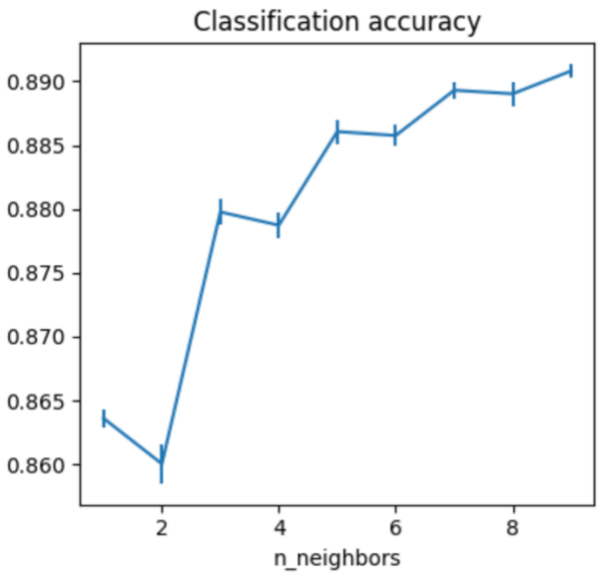

Fig. 3.1.2 Accuracy of test data with respect to the n neighbors (K)

## 3.2 Artificial Neural Networks for Galaxy Classification

During the past ten years, astronomical data has been gathered by a new class of large telescopes capable of collecting vast quantities of data; the volume of data collected from an entire survey ten years ago can now be gathered within one night with a single telescope. The fifth Sloan Digital Sky Survey (SDSS-V) is one of these new generation telescopes, which will be launched in 2020 and slated to collect optical and infrared spectra for more than six million objects over the course of its five-year life span. Although Artificial Neural Networks (ANN) are currently in boom to deal with a plethora of the data. For a vast amount of astronomical data, ANNs are increasingly popular [3] as a major technique of classifying the data. An ANN is one that can deal with large amounts of computationally intensive formulas and statistical techniques with ease and a high level of accuracy. As a result of human intervention, there is always the possibility of making mistakes when doing calculations or running into trouble when dealing with time-consuming findings.

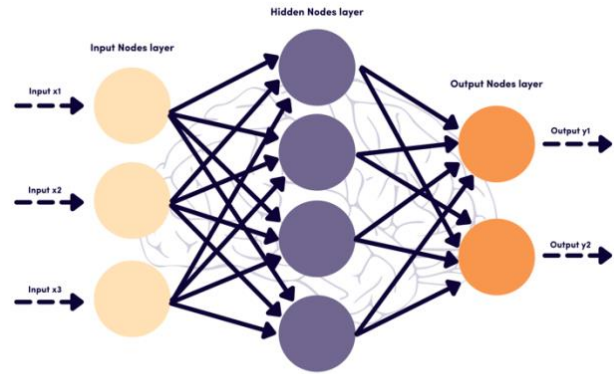

Fig. 3.2.1 Artificial Neural Network

A typical ANN comprises three primary layers: input, hidden, and output, with a collection of fundamental units called neurons in each layer. A neuron receives an input, processes it using an aggregation function, and then produces an output so that the process can be continued by the following neuron in the structure. An ANN has two flows: a *forward pass/propagation flow* and a *backward pass/propagation flow*. Forward propagation involves moving through the hidden layer from the input layer (left) to the output layer (right). With backward propagation, we pass the hidden layers in between as we move from the output layer (on the right) to the input layer (on the left). Shape_of_galaxy is a multi-class target variable that we have included in our network. There are three neurons in the output, and each is classified as Spiral, Elliptical, or Uncertain [7]. We have used 10 characteristics to create 10 neurons for the input layer. Each layer also has an *activation function [6]*, the activation function relates to the forward propagation of calculation and storage of intermediate variables through the network. This function forwards the output while ensuring that values are kept within a range that is acceptable and useful for following layers [4].

$$z = W^{(1)}x$$

Where $W^{(1)} \in \mathbb{R}^{h \times d}$ is the weight parameter of the hidden layer. After running the intermediate variable $z \in \mathbb{R}^h$ through the activation function $\phi$ we obtain our hidden activation vector of length h.

$$h = \phi(z).$$

The hidden layer output h is also an intermediate variable. Assuming that the parameters of the output layer only possess a weight of $W^{(2)} \in \mathbb{R}^{q \times h}$, we can obtain an output layer variable with a vector of length q :

$$o = W^{(2)}h$$

Assuming that the loss function is l and the example label is, we can then calculate the loss term for a single data example,

$$L = l(o, y)$$

According to the definition of ℓ2 regularization that we will introduce later, given the hyperparameter λ, the regularization term is:

$$s = \lambda/2(\|W^{(1)}\|^2_F + \|W^{(2)}\|^2_F),$$

where the Frobenius norm of the matrix is simply the ℓ2 norm applied after flattening the matrix into a vector. Finally, the model's regularized loss on a given data example is:

$$J = L + s$$

We refer to J as the *objective function*.
For backward propagation, assume that we have functions $Y = f(X)$ and $Z = g(Y)$, in which the input and the output X, Y, Z are tensors of arbitrary shapes. By using the chain rule, we can compute the derivative of Z with respect to X via

$$\frac{\partial Z}{\partial X} = \text{prod}\left(\frac{\partial Z}{\partial Y}, \frac{\partial Y}{\partial X}\right)$$

Here we use the prod operator to multiply its arguments after the necessary operations, such as transposition and swapping input positions, have been carried out.

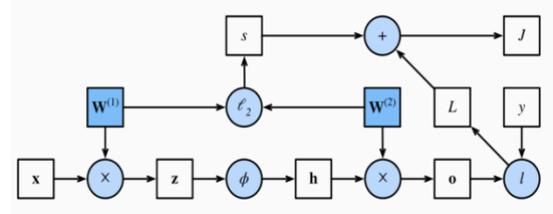

Fig. 3.2.2 Computational graph of forward propagation

A. *Algorithm:*

7. Input: The neuron receives inputs from neurons in the previous layer.
8. Weight: Each input is multiplied by a weight value (w1, w2, w3) that is specific to the neuron.
9. Summation: The weighted inputs are summed up.
10. Bias: A bias value may be added to the sum.
11. Activation function: The result of the summation is passed through an activation function to produce the output of the neuron.
12. Output: The output of the neuron is passed as an input to neurons in the next layer.

B. *Implementation:*

With the help of the train_test_split() method from the sklearn.model_selection module, the dataset was split into training and testing sets. For the sake of repeatability, the data was divided into training and testing groups in proportions of 70:30 each, with a random state of 17.
Before being converted into categorical format, the training and testing labels were preprocessed by using the to_categorical() function from the tensorflow.keras.utils module to fill in any missing values with a value of 3.0. To confirm that the transformation was accurate, the nunique() method was used to check the number of distinct classes in the training set.

Utilizing the Sequential() function from the tensorflow.keras.models module, the neural network architecture was created. Three dense layers made up the network; the final output layer used a softmax activation function, while the preceding two layers used ReLU activation functions. The number of features in the training data was used to establish the input shape for the first layer. The optimizer was set to "adam" and the loss function to "categorical_crossentropy" when the model was built using the compile() method. To track the model's performance during training, the metrics were set to "accuracy."

A hyperparameter tuning process was performed using RandomizedSearchCV to search for the best combination of hyperparameters for a deep neural network model. The model was built using the KerasRegressor wrapper from the TensorFlow library. The search space was defined using a dictionary of hyperparameters including the number of hidden units, activation function, and optimizer. The best model was selected based on the mean squared error metric evaluated using a 3-fold cross-validation. The model was trained on the training data using the best hyperparameters identified from the search. The accuracy and loss metrics were recorded during training to assess the model's performance. The best model was selected based on the hyperparameters that produced the best performance on the validation set.

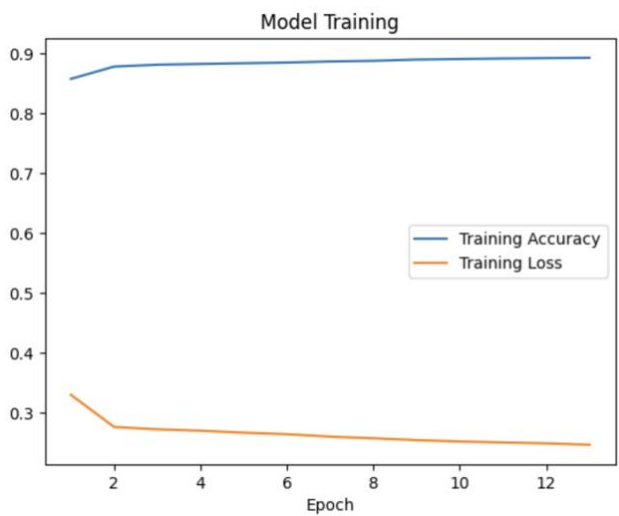

Fig 3.2.3 Accuracy vs loss graph for ANN model

*C. Outcome:*

In this study, a deep learning model was developed to classify numeric data of different types of galaxies [10]. The model was trained and tested on a dataset of 60000 records, with an accuracy score of 0.89 achieved on the test set. The accuracy score was calculated using the scikit-learn library and indicates the proportion of correct predictions made by the model.

To visualize the accuracy of the model, a scatter plot was generated comparing the actual values of the test set with the predicted values. The plot shows a strong positive correlation between the actual and predicted values, indicating that the model is performing well in predicting the class labels of the test set.

Overall, the results of this study suggest that deep learning models can be effective in classifying galaxy data. The model developed in this study achieved a high level of accuracy, indicating its potential use in real-world applications. Further studies could investigate the performance of the model on a larger and more diverse dataset, as well as exploring other types of deep learning models for classification.

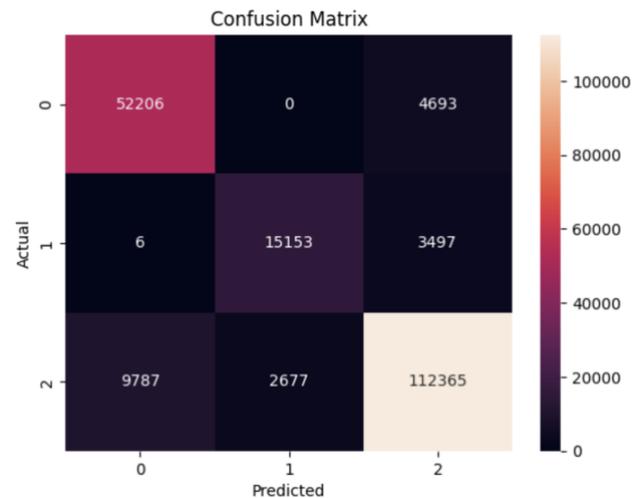

Fig 3.2.4 Actual vs Predicted values heatmap.

## 3. Conclusion

In this study, we compared the performance of two different machine learning algorithms, K-Nearest

Neighbors (KNN) and Artificial Neural Networks (ANN), for classifying galaxies based on kinematic and numerical data. Our results indicate that both algorithms achieved the same accuracy of 89% in classifying the galaxies. This suggests that both algorithms can be equally effective in accurately classifying galaxies, and the choice of algorithm may depend on other factors such as data availability, computational resources, and interpretability of the results. Overall, the findings of this study demonstrate the potential of machine learning approaches in accurately classifying galaxies and provide a foundation for further research in this area.

## *4. Future Enhancements*

The galaxy classification model using numerical features and K-Nearest Neighbor (KNN) and Artificial Neural Networks (ANN) has also shown promising results in accurately classifying galaxies into different types. However, there are several potential future enhancements that could be explored to improve the model's performance and applicability:

1. **Data augmentation**: To improve the performance of the model, we can consider data augmentation techniques such as rotation, flipping, and scaling to increase the amount of data available for training.
2. **Feature engineering**: We can explore additional features that may be relevant for classifying galaxies, such as the presence of spiral arms or the distribution of star formation. Additionally, we can consider using more advanced feature extraction techniques such as principal component analysis (PCA) or t-distributed stochastic neighbor embedding (t-SNE)[1].
3. **Ensemble methods**: Combining multiple models, such as KNN and ANN, into an ensemble can improve the overall classification accuracy. We can explore different ensemble methods, such as bagging or boosting, to improve the model's performance.
4. **Transfer learning**: Transfer learning techniques can be used to leverage pre-trained models on similar tasks or data. For example, we can use pre-trained models trained on other image classification tasks or data to improve the performance of our galaxy classification model.
5. **Explainability:** To improve the interpretability and explainability of the model, we can use techniques such as saliency maps or gradient-weighted class activation mapping (Grad-CAM) to visualize which features of the galaxies are most important for the classification task.